\documentclass[10pt,twocolumn,letterpaper]{article}

\usepackage{cvpr}              

\usepackage{graphicx}
\usepackage{amsmath}
\usepackage{amssymb}
\usepackage{booktabs}
\usepackage{times}
\usepackage{epsfig}
\usepackage{graphicx}
\usepackage{amsmath}
\usepackage{amssymb}
\usepackage{enumitem}
\usepackage{wrapfig}
\usepackage{tablefootnote}
\usepackage{makecell}
\usepackage{booktabs}
\usepackage{siunitx}

\newcommand{\rr}{\mathbf{r}}
\newcommand{\ttt}{\mathbf{t}}
\newcommand{\WW}{\mathbf{W}}

\usepackage[pagebackref,breaklinks,colorlinks]{hyperref}

\usepackage[capitalize]{cleveref}
\crefname{section}{Sec.}{Secs.}
\Crefname{section}{Section}{Sections}
\Crefname{table}{Table}{Tables}
\crefname{table}{Tab.}{Tabs.}

\def\cvprPaperID{18} 
\def\confName{CVPR}
\def\confYear{2022}

\begin{document}

\title{BlazePose GHUM Holistic: Real-time 3D Human Landmarks and Pose Estimation}

\author{
Ivan Grishchenko \quad Valentin Bazarevsky \quad Andrei Zanfir \quad Eduard Gabriel Bazavan \quad Mihai Zanfir\\
Richard Yee \quad Karthik Raveendran \quad Matsvei Zhdanovich \quad Matthias Grundmann \quad Cristian Sminchisescu\\
Google\\
1600 Amphitheatre Pkwy, Mountain View, CA 94043, USA\\
{\tt\footnotesize \{igrishchenko, valik, andreiz, egbazavan, mihaiz, yeer, krav, matvey, grundman, sminchisescu\}@google.com}\\
}
\maketitle

\begin{abstract}

We present BlazePose GHUM Holistic, a lightweight neural network pipeline for 3D human body landmarks and pose estimation, specifically tailored to real-time on-device inference. BlazePose GHUM Holistic enables motion capture from a single RGB image including avatar control, fitness tracking and AR/VR effects. Our main contributions include i) a novel method for 3D ground truth data acquisition, ii) updated 3D body tracking with additional hand landmarks and iii) full body pose estimation from a monocular image.

\end{abstract}

\section{Introduction}


Accurate real-time inference of the human body skeleton enables a variety of applications ranging from motion capture to interactive video games. Running on a consumer's phone or laptop without relying on dedicated sensors (\eg IR on the Kinect \cite{Kinect}) would vastly democratize the technology for non-professional users. Over the past decade, there have been numerous advances in estimating 3D landmarks on the human body \cite{Guler_2019_CVPR} or volumetric representations \cite{BodyNet}. The majority of these are either too computational expensive to be run on mobile devices, require a specialized lab setup (\eg multi-camera) or lack sufficient detail \wrt body topology (\eg no fingers). To overcome these limitations, we created BlazePose GHUM Holistic, a lightweight neural network pipeline that predicts 3D landmarks and pose of the human body on-device, including hands, from a single monocular image, runs in real-time at 15 FPS on most modern mobile phones and browsers and is available to developers and creators via MediaPipe.

Our approach handles three issues that we identified with current interactive motion capture solutions. First, we address the challenge of acquiring \textit{diverse} 3D ground truth of the human body. We introduce a novel approach that is based on fitting a statistical 3D human model GHUM\cite{ghum2020} to a diverse set of 2D annotations. To further improve accuracy, we propose to use depth ordering annotations as supervision during fitting.

Second, the standard topology for on-device body landmarks prediction does not include hands and fingers, \eg 33 landmarks by BlazePose\cite{blazepose2020} and PoseNet \cite{PoseNet}. This impedes the building of a unified motion capture system for the entire body. To overcome this limitation, we use a spatial transformer \cite{SpatialTransformers} approach to crop high-res hand regions from the original image seeded by BlazePose's palm prediction as a prior. Then we run a retrained hand tracking model \cite{Zhang20, sung2021device} to predict 21 3D hand landmarks for each hand in a single feed-forward pass.

\begin{figure}
  \centering
  \includegraphics[width=1.0\linewidth]{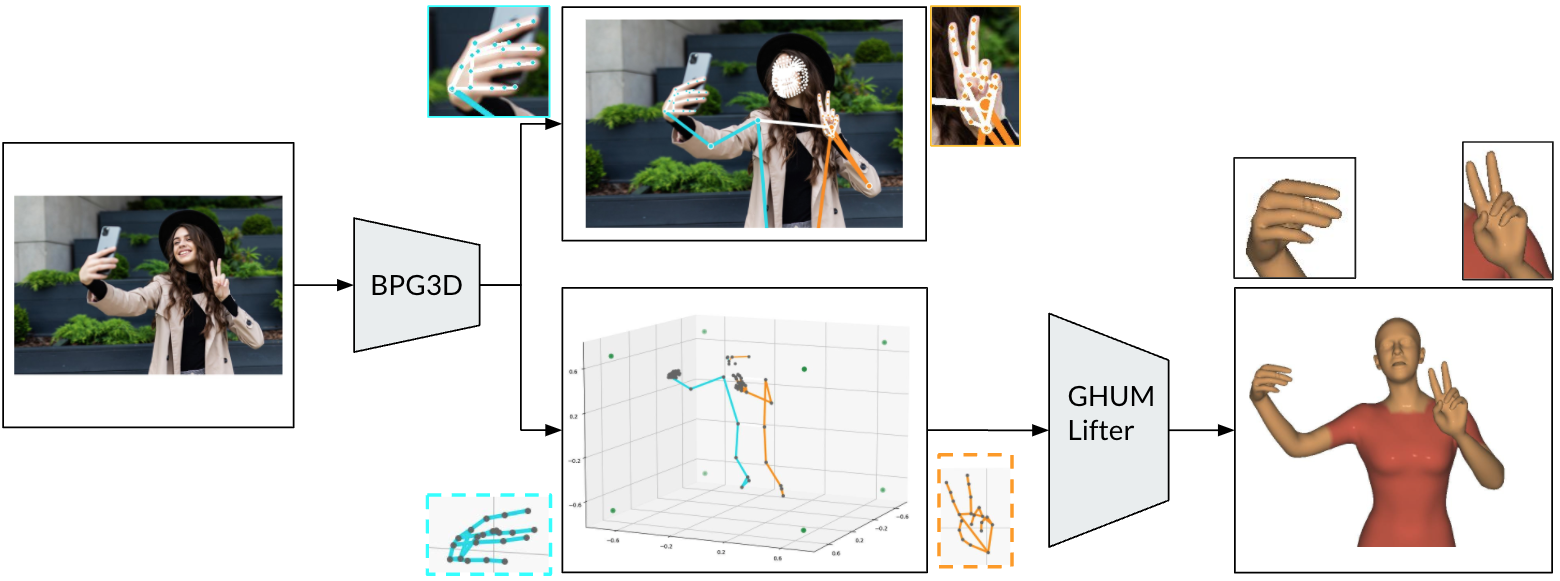}
  \caption{Overview of our BlazePose GHUM Holistic pipeline. From a single RGB input image, we predict 2D and 3D landmarks for body and hands in one feedforward pass. Full 3D pose and shape information is further inferred via a GHUM\cite{ghum2020} lifter.}
  \label{fig:blazeposeghum3d_pipeline}
\end{figure}

Finally, we tackle the issue of moving beyond accurate 3D representations of the human body towards high-level semantic understanding and mapping to enable expressive use cases like 3D avatars. To this end, we present a lightweight model that predicts the full body and hand pose from 3D landmarks represented as joint rotations of a 3D human model. We employ a statistical 3D human model called  GHUM\cite{ghum2020} that additionally acts as a pose prior constraining the state of predictions to plausible and realistic movements.

\section{3D body landmarks}

\begin{figure}
  \centering
  \includegraphics[width=1.0\linewidth]{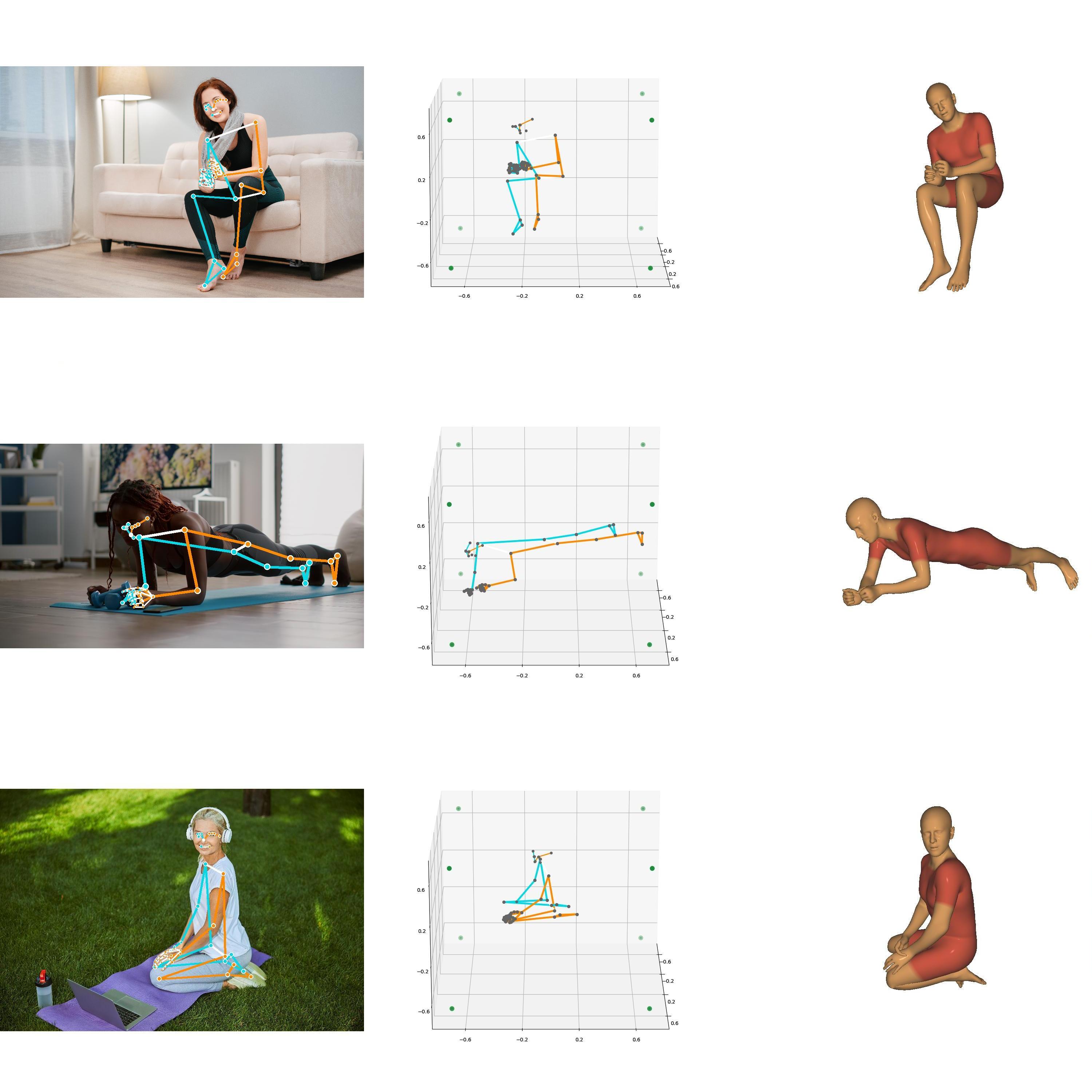}
  \caption{Sample results of BlazePose GHUM Holistic. For an input image our pipeline predicts real-time on-device 2D and 3D landmarks for body and hands (first and second columns) as well as GHUM\cite{ghum2020} pose parameters (third column). Our methods work for challenging poses, various backgrounds and diverse appearances.}
  \label{fig:main_example}
\end{figure}

The key challenge to build the 3D part of our pose model is obtaining realistic, \textit{in-the-wild} 3D data. In contrast to 2D, which can be obtained via human annotation, accurate manual 3D annotation is a uniquely challenging task. It requires either a lab setup\cite{li2021learn, h36m_pami}, specialised hardware with depth sensors for 3D scans \cite{ghum2020} or building a synthetic dataset\cite{bazavan2021hspace,simpose2020} which has an inherent domain gap to real world data. Each of these approaches introduces additional challenges to preserve a good level of human and environment diversity as present in real-world pictures.

\begin{figure*}
\begin{center}
\end{center}
   \includegraphics[width=\linewidth]{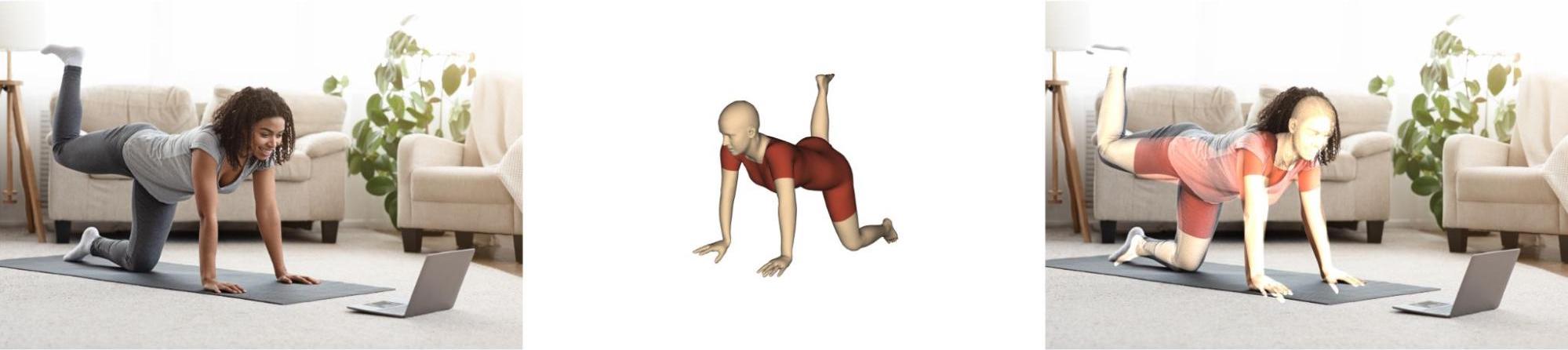}
   \caption{Sample GHUM fitting for an input image. Left: original image, middle: 3D GHUM reconstruction (different viewpoint), right: blended result projected on top of the original image.}
   \label{fig:ghum_fitting}
\end{figure*}

\paragraph{3D data acquisition}
Our approach is based on a statistical 3D human body model GHUM\cite{ghum2020}, which is built using a large corpus of human shapes and motions. To obtain 3D human body pose ground truth, we fit the GHUM model to our existing 2D pose dataset, which covers various domains (yoga/fitness/dance), surroundings (indoor/outdoor), devices (mobile/laptop).  In doing so, we obtain real world 3D keypoint coordinates in metric space (see fig. \ref{fig:ghum_fitting}). During the fitting process the shape and the pose variables of GHUM were optimized such that the reconstructed model aligns with the underlying image evidence. This includes 2D keypoint and silhouette semantic segmentation alignment as well as shape and pose regularization terms (check HUND\cite{zanfir2020neural}, THUNDR\cite{Zanfir_2021_ICCV}).


Due to the nature of 3D to 2D projection fitting can result in several realistic 3D body poses for the given 2D annotation (\ie with the same X and Y but different Z). To minimize this ambiguity we asked annotators to provide depth order between pose skeleton edges where they are certain, similarly to Ordinal Depth Supervision approach\cite{pavlakos2018ordinal}. This task proved to be easy showing high consistency between annotators (98\% on cross-validation) and helped to reduce the depth ordering errors during fitting from 25\% to 3\%.

\paragraph{Model}
BlazePose GHUM Holistic utilizes a two-step detector-tracker approach where the tracker operates on a cropped region-of-interest containing the human within the original image. Thus the model is trained to predict 3D body pose in relative coordinates of a metric space with origin in the subject's hips center.

\paragraph{Experiments}
To evaluate the quality of our models against other well-performing publicly available solutions, we use yoga domain, as one of the most challenging in body poses articulations. Each image contains only a single person located 2-4 meters from the camera. To be consistent with other solutions, we perform evaluation only for 17 keypoints from COCO topology. As shown in \cref{tbl:3d_leaderboard}, our approach outperforms that of leading commercial and academic solutions. We train a variety of different models (Heavy, Full, Lite) that provide various levels of trade-off between accuracy and on-device inference speed, see \cref{tbl:bp_inference}.

\begin{table}[ht]
    \begin{center}
    \begin{tabular}[t]{|l|r|r|r|}
    \hline
    Model & 2D & 3D \\
     & mAP & mae (mm) \\ 
    \hline
    \textbf{BlazePose Heavy} & $\mathbf{68.1}$ & $\mathbf{36}$ \\
    \hline
    BlazePose Full & 62.6 & 39 \\
    \hline
    BlazePose Lite & 45.0 & 45 \\
    \hline
    AlphaPose ResNet50 & 63.4 & N/A \\
    \hline
    Apple Vision & 32.8 & N/A \\
    \hline
    \end{tabular}
    \end{center}
\caption{Comparison for single person pose tracking methods.}
\label{tbl:3d_leaderboard}
\end{table}

\begin{table}[ht]
    \begin{center}
    \begin{tabular}[t]{|l|r|r|r|}
    \hline
    Device & BlazePose & BlazePose & BlazePose \\
     & Lite, ms & Full, ms & Heavy, ms \\ 
    \hline
    In-Browser\tablefootnote{Google Chrome on MacBook Pro (15-inch 2017)} & 13 & 15 & 29 \\
    \hline
    Pixel 4 CPU \tablefootnote{Single Core via XNNPACK backend} & 25 & 40 & 147 \\
    \hline
    Pixel 4 GPU & 8 & 9 & 22 \\
    \hline
    Desktop \tablefootnote{Intel i9-10900K. Nvidia GTX 1070 GPU} & 7 & 8 & 10 \\
    \hline
    \end{tabular}
    \end{center}
\caption{Inference speed across various devices}
\label{tbl:bp_inference}
\end{table}

\section{Hand landmarks for holistic human pose}

Holistic human pose estimation requires accurate tracking of hands in addition to the body. To include hand landmarks in the BlazePose topology, we had to solve two main issues. First, the BlazePose model's input resolution of 256x256 is insufficient to capture hand details. Second, we need to ensure that hands prediction is spatially invariant for left and right hands to guarantee the same level of accuracy as leading on-device hand tracking models \cite{sung2021device}. Using a single model would require us to a) increase the input resolution which in turn would slow down model inference and b) balance the loss for small and big body parts a single pixel error is more significant for fingers, than for example the hip landmark. Instead, we opted for an alternative approach that uses a separate hand prediction model\cite{sung2021device}that is inferred on high resolution crops obtained from ROIs provided by BlazePose (\ie re-cropping).

\paragraph{Pipeline and re-crop model}
The four palm landmarks produced by BlazePose give us a rough estimate of the hand region. But this is insufficient to use as input for the subsequent hand landmark model \cite{sung2021device} as it was trained on more accurate crops obtained from all 21 hand landmarks with only slight transformation augmentations. One solution would be to train with more aggressive transformation augmentations but this significantly reduces accuracy of the model due to its limited capacity targeting real-time on-device inference. To close this gap, we trained a re-crop model that takes raw crop from BlazePose output and refines it on a higher resolution to a level acceptable for the subsequent hand landmark model. Using BlazePose as a prior for hand locations also helps us to untangle hands and body parts of different people.

\paragraph{Experiments}
We compare the hand landmarks quality of a standalone hand tracker \cite{sung2021device} with our novel re-crop approach, see \cref{tbl:recrop_error}. Our pipeline produces a better hand landmark quality. We postulate this is due to a more accurate crop on the current frame, whereas the approach of \cite{sung2021device} uses a one frame delay.

\begin{table}[!htbp]
    \begin{center}
    \begin{tabular}[t]{|l|r|r|r|}
    \hline
     Pipeline & 2D & 3D \\
     & mnae & mae (mm))\\
    \hline
    Tracking pipeline (baseline) & 9.8\% & 20 \\
    \hline
    Pipeline without re-crops & 11.8\% & 27 \\
    \hline
    \textbf{Pipeline with re-crops} & $\mathbf{9.7}$\% & $\mathbf{18}$ \\
    \hline
    \end{tabular}
    \end{center}
\caption{Hand prediction quality.The mean error per hand (MEH) is normalized by the hand size.}
\label{tbl:recrop_error}
\end{table}

\section{Body pose estimation}

To enable expressive use cases like 3D avatars, we must go beyond inferring 3D coordinates and obtain a high-level semantic understanding via joint rotations that can be used to drive rigged characters. To this end, we obtain the 3d pose and shape GHUM mesh from the set of our previously 3d landmarks inferred directly from images.

\paragraph{Statistical human model.} We use the recently introduced statistical 3d human body model GHUM \cite{ghum2020}, to represent the pose and shape of the human body. The model has been trained end-to-end, in a deep learning framework, using a large corpus of human shapes and motions. 
The model has generative body shape and facial expressions $\beta = \left( \beta_b, \beta_f \right) $ represented using deep variational auto-encoders and generative pose $\theta = \left( \theta_{b}, \theta_{lh}, \theta_{rh} \right)$ for the body, left and right hands, respectively, represented as normalizing flows \cite{zanfir2020weakly}. The pelvis translation and rotation are controlled separately, and represented by a 6d rotation representation \cite{zhou2018continuity} $\rr\in\mathbb{R}^{6\times1}$ and a translation vector $\ttt\in\mathbb{R}^{3\times1}$ w.r.t the origin $(0, 0, 0)$. The mesh consists of of $N_v = \num{10168}$ vertices and $N_t = \num{20332}$ triangles. To pose the mesh, we apply the GHUM network $\mathbf{V}(\theta_{b}, \beta_{b}, \rr, \ttt) \in \mathbb{R}^{N_v \times 3}$ to obtain the posed vertices. We omit the facial expressions, as we here focus on main body, hand poses and shape. We also drop the $b$ subscript for convenience.

\paragraph{Lifter.}
The BlazePose GHUM Holistic network outputs 33 body landmarks, and 21 landmarks for each hand, in a root-centered 3D camera coordinate system. In order to increase the expressivity of the outputs, without sacrifing performance, we propose a sample-and-train methodology, based on a novel \textit{GHUM Lifter} neural network. Our neural network takes as input the concatenated 3D body and hands landmarks and outputs GHUM mesh parameters.

\paragraph{MLPMixer.}
At the core of our lifter lies an MLP-Mixer architecture \cite{tolstikhin2021mlp}. 
Our mixer takes as input a sequence of S 3D keypoints, each one projected to a desired
hidden dimension C, with the same projection matrix. The mixer
consists of multiple layers of identical size, and each layer consists of two MLP blocks. The first one is the token-mixing MLP: it acts on columns of the mixer input and is shared across all columns. The second one is the channel-mixing MLP: it
acts on rows of the token-mixing MLP and is shared across all rows. Each MLP block contains two fully-connected layers and a nonlinearity applied independently to each row of its input data tensor. 

\paragraph{Full model training.}
We use an MLPMixer architecture which takes as input $S=\num{75}$ tokens (\ie 2 x 21 hand 3D landmarks and 33 body 3D landmarks) and transforms them to produce GHUM state parameters $\rr, \ttt, \beta, \theta$. Given generative codes for pose and shape $\theta, \beta \in \mathcal{N}(\mathbf{0}; I)$, $\rr$ drawn from the Haar distribution on $SO(3)$, and $\ttt$ uniformly sampled from a $(\num{-0.1} \ldots \num{-0.1}) \times (\num{-0.1} \ldots \num{0.1}) \times (\num{-0.1} \ldots \num{-0.1})$ meters box, we produce a posed GHUM  sample mesh $\mathbf{V}(\theta, \beta, \rr, \ttt)$. The associated $S$ 3D landmarks can be retrieved by a simple (fixed) linear regression matrix $\WW \in \mathbb{R}^{N_v \times S} $, such that $\mathbf{X} = \WW\mathbf{V}(\theta, \beta, \rr, \ttt)$. In our experiments, we noticed that injecting noise at this point, \ie $\mathbf{X} + \mathcal{N}(\mathbf{0}; \epsilon I)$, supports the more accurate retrieval of the full mesh given real. We also experimented with Transformer \cite{vaswani2017attention} or simple MLP architectures. The latter failed to converge (which was also experienced by \cite{Zanfir_2021_ICCV}), while the former had a similar performance, but required more memory and it was harder to deploy on mobile devices.  


\paragraph{Experiments.} In order to validate our GHUM lifter architecture we ran experiments on a held out test set of 10,000 in the wild images with very challenging poses (yoga, fitness, dancing) containing GHUM fits which were curated for any errors. We compared with various SOTA methods for 3D pose and shape estimation and observe significant improvements for our method even though it runs one order of magnitude faster. Results are reported in \cref{tbl:comparison_oi_test}.


\begin{table}[!htbp]
    \small
    \centering
    \begin{tabular}[t]{|l|r|r|}
    \hline
    Method & {MPJPE-PA} & {MPJPE} \\
    \hline
    SPIN \cite{kolotouros2019learning} & $\num{101}$ & $\num{139.5}$ \\
    \hline
    HUND \cite{zanfir2020neural} (WS) & $\num{106.6}$ & $\num{156}$ \\
    \hline
    THUNDR \cite{Zanfir_2021_ICCV} (WS) & $\num{97.5}$ & $\num{138}$ \\
    \hline
    \textbf{BlazePose GHUM Holistic} & $\mathbf{78}$ & $\textbf{121}$ \\
    \hline
    \end{tabular}
    \caption{\small Mean per joint positional error with Procrustes alignment (MPJPE-PA) and without Procrustes alignment (MPJPE) on a held out test set containing in the wild images  with very complex poses (yoga, fitness, dancing}
\label{tbl:comparison_oi_test}
\end{table}

\section{Applications and conclusion}

\begin{figure}[h]
  \begin{center}
   \includegraphics[width=0.9\linewidth]{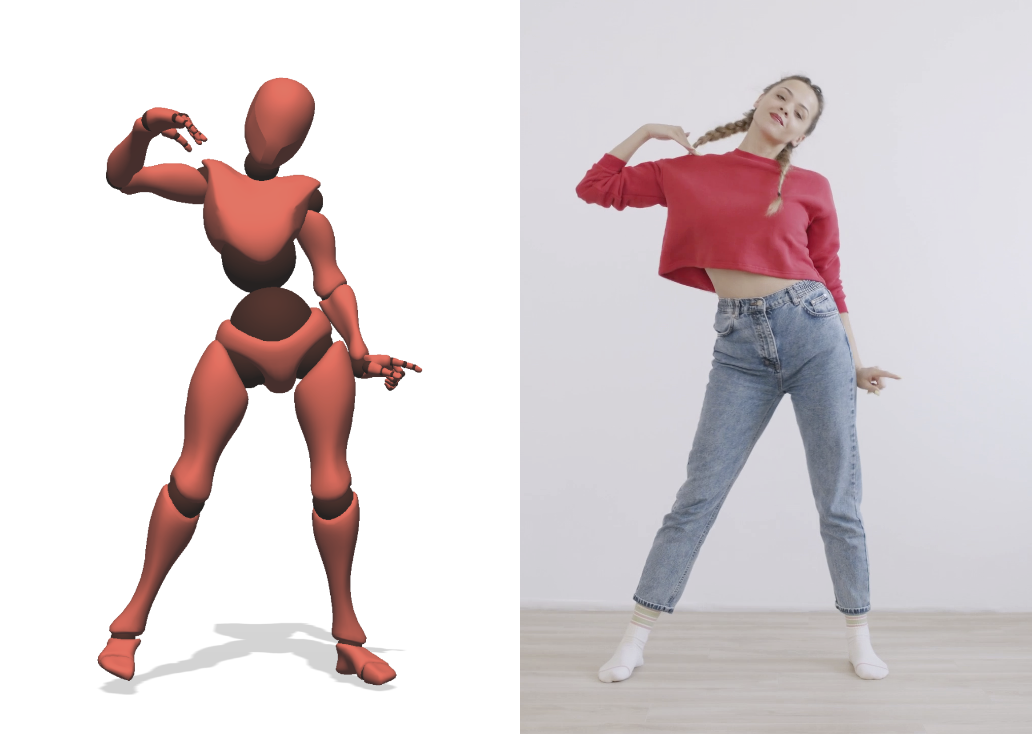}
  \end{center}
  \caption{Open-sourced avatar demo in MediaPipe.}
  \label{fig:avatar_demo}
\end{figure}

BlazePose GHUM Holistic provides a simple and meaningful representation of the human body that can be used for multiple applications right out of the box. 3D landmarks enable us to move beyond 2D space to a real world coordinate system. Pose estimation gives us high level interpretation of 3D landmarks as well as extra hand points to add fine grain details when needed (\eg gesture detection). It enables motion capture for avatar control, repetition counting and posture correction for fitness and sports, as well as 3D effects for AR/VR.

To showcase BlazePose GHUM Holistic we created an open-source avatar demo (see \cref{fig:avatar_demo}) in MediaPipe (\url{https://mediapipe.dev}). The demo is available for web browsers and allows to control body and hands of a standard Mixamo avatar at 15 FPS (MacBook Pro 15-inch 2017).

In the future, we plan on further improving the model that predicts 3D landmarks as well as capturing reliable facial expressions to add them to BlazePose GHUM Holistic.

{\small
\bibliographystyle{ieee_fullname}
\bibliography{egbib}

\begin{thebibliography}{10}\itemsep=-1pt

\bibitem{Kinect}
Azure kinect body tracking joints.
\newblock \\\url{https://docs.microsoft.com/en-us/azure/kinect-dk/body-joints}.

\bibitem{Guler_2019_CVPR}
Riza Alp~Guler and Iasonas Kokkinos.
\newblock Holopose: Holistic 3d human reconstruction in-the-wild.
\newblock In {\em CVPR}, 2019.

\bibitem{blazepose2020}
Valentin Bazarevsky, Ivan Grishchenko, Karthik Raveendran, Tyler Zhu, Fan
  Zhang, and Matthias Grundmann.
\newblock Blazepose: On-device real-time body pose tracking, 2020.

\bibitem{bazavan2021hspace}
Eduard~Gabriel Bazavan, Andrei Zanfir, Mihai Zanfir, William~T. Freeman, Rahul
  Sukthankar, and Cristian Sminchisescu.
\newblock Hspace: Synthetic parametric humans animated in complex environments,
  2021.

\bibitem{Zhang20}
Andrey~Vakunov Fan~Zhang, Valentin~Bazarevsky, George Sung, Chuo-Ling Chang,
  Matthias Grundmann, and Andrei Tkachenka.
\newblock Mediapipe hands: On-device real-time hand tracking.
\newblock In {\em CVPR}, 2020.

\bibitem{h36m_pami}
Catalin Ionescu, Dragos Papava, Vlad Olaru, and Cristian Sminchisescu.
\newblock Human3.6m: Large scale datasets and predictive methods for 3d human
  sensing in natural environments.
\newblock {\em IEEE Transactions on Pattern Analysis and Machine Intelligence},
  36(7):1325--1339, jul 2014.

\bibitem{SpatialTransformers}
Max Jaderberg, Karen Simonyan, Andrew Zisserman, and Koray Kavukcuoglu.
\newblock Spatial transformer networks, 2015.

\bibitem{PoseNet}
Alex Kendall, Matthew Grimes, and Roberto Cipolla.
\newblock Posenet: A convolutional network for real-time 6-dof camera
  relocalization, 2015.

\bibitem{kolotouros2019learning}
Nikos Kolotouros, Georgios Pavlakos, Michael~J Black, and Kostas Daniilidis.
\newblock Learning to reconstruct 3d human pose and shape via model-fitting in
  the loop.
\newblock In {\em ICCV}, 2019.

\bibitem{li2021learn}
Ruilong Li, Shan Yang, David~A. Ross, and Angjoo Kanazawa.
\newblock Learn to dance with aist++: Music conditioned 3d dance generation,
  2021.

\bibitem{pavlakos2018ordinal}
Georgios Pavlakos, Xiaowei Zhou, and Kostas Daniilidis.
\newblock Ordinal depth supervision for 3{D} human pose estimation.
\newblock In {\em Computer Vision and Pattern Recognition (CVPR)}, 2018.

\bibitem{sung2021device}
George Sung, Kanstantsin Sokal, Esha Uboweja, Valentin Bazarevsky, Jonathan
  Baccash, Eduard~Gabriel Bazavan, Chuo-Ling Chang, and Matthias Grundmann.
\newblock On-device real-time hand gesture recognition.
\newblock 2021.

\bibitem{tolstikhin2021mlp}
Ilya Tolstikhin, Neil Houlsby, Alexander Kolesnikov, Lucas Beyer, Xiaohua Zhai,
  Thomas Unterthiner, Jessica Yung, Andreas Steiner, Daniel Keysers, Jakob
  Uszkoreit, et~al.
\newblock Mlp-mixer: An all-mlp architecture for vision.
\newblock 2021.

\bibitem{BodyNet}
Gül Varol, Duygu Ceylan, Bryan Russell, Jimei Yang, Ersin Yumer, Ivan Laptev,
  and Cordelia Schmid.
\newblock Bodynet: Volumetric inference of 3d human body shapes, 2018.

\bibitem{vaswani2017attention}
Ashish Vaswani, Noam Shazeer, Niki Parmar, Jakob Uszkoreit, Llion Jones,
  Aidan~N Gomez, {\L}ukasz Kaiser, and Illia Polosukhin.
\newblock Attention is all you need.
\newblock {\em Advances in neural information processing systems}, 30, 2017.

\bibitem{ghum2020}
Hongyi Xu, Eduard~Gabriel Bazavan, Andrei Zanfir, Bill Freeman, Rahul
  Sukthankar, and Cristian Sminchisescu.
\newblock {GHUM} \& {GHUML}: Generative {3D} human shape and articulated pose
  models.
\newblock {\em CVPR}, 2020.

\bibitem{zanfir2020weakly}
Andrei Zanfir, Eduard~Gabriel Bazavan, Hongyi Xu, Bill Freeman, Rahul
  Sukthankar, and Cristian Sminchisescu.
\newblock Weakly supervised 3d human pose and shape reconstruction with
  normalizing flows.
\newblock {\em ECCV}, 2020.

\bibitem{zanfir2020neural}
Andrei Zanfir, Eduard~Gabriel Bazavan, Mihai Zanfir, William~T Freeman, Rahul
  Sukthankar, and Cristian Sminchisescu.
\newblock Neural descent for visual 3d human pose and shape.
\newblock 2020.

\bibitem{Zanfir_2021_ICCV}
Mihai Zanfir, Andrei Zanfir, Eduard~Gabriel Bazavan, William~T. Freeman, Rahul
  Sukthankar, and Cristian Sminchisescu.
\newblock Thundr: Transformer-based 3d human reconstruction with markers.
\newblock In {\em ICCV}, 2021.

\bibitem{zhou2018continuity}
Yi Zhou, Connelly Barnes, Jingwan Lu, Jimei Yang, and Hao Li.
\newblock On the continuity of rotation representations in neural networks.
\newblock 2018.

\bibitem{simpose2020}
Tyler Zhu, Per Karlsson, and Christoph Bregler.
\newblock Simpose: Effectively learning densepose and surface normals of people
  from simulated data, 2020.

\end{thebibliography}
}

\end{document}